\documentclass[10pt,twocolumn,letterpaper]{article}

\usepackage[pagenumbers]{cvpr}

\usepackage{colortbl}
\usepackage{multirow}

\definecolor{cvprblue}{rgb}{0.21,0.49,0.74}
\usepackage[pagebackref,breaklinks,colorlinks,allcolors=cvprblue]{hyperref}

\title{HDBox3D: Box-Guided High-Resolution Processing\\for Efficient Interactive 3D Segmentation}

\author{
Yuhi Chiba \qquad Hiroshi Ishikawa\\
Department of Computer Science and Engineering\\
Waseda University, Tokyo, Japan
}

\begin{document}
\maketitle
\begin{abstract}
Interactive 3D point cloud instance segmentation enables users to iteratively refine object masks through corrective feedback.
Existing methods repeatedly process the entire scene at every interaction, even though only a limited region around the target object is typically relevant.
To address this inefficiency, we introduce \textbf{FastBox3D}, which constructs a localized interaction region from the previous mask and the latest feedback, substantially improving efficiency while maintaining segmentation accuracy.
Building on this computational saving, we further propose \textbf{HDBox3D}, which extracts higher-resolution features within the localized region to improve segmentation quality while retaining much of the efficiency gain over the base model.
In addition, we introduce \textbf{Voxel Size Alignment (VSA)} to mitigate the performance degradation due to the differences in point density between the training and target domains.
VSA adjusts the voxel size according to the relative point spacing of the two domains, producing voxel representations at inference time that better match those observed during training.
Overall, our approach achieves state-of-the-art or competitive performance across all four evaluated datasets: ScanNet, S3DIS, KITTI-360, and ScanNet++.
\end{abstract}
\section{Introduction}
\label{sec:intro}

3D point cloud instance segmentation is fundamental to 3D scene understanding and supports a wide range of applications such as robotics and autonomous driving. However, obtaining dense instance annotations for large point clouds requires substantial manual effort~\cite{dai2017scannet,liao2022kitti360}. Interactive 3D point cloud segmentation alleviates this burden by allowing users to iteratively correct segmentation errors by providing positive and negative hints through mouse clicks. Since the proposal of InterObject3D~\cite{kontogianni2023interobject3d}, various ``click-based'' methods have emerged to enhance segmentation accuracy and interaction efficiency~\cite{wen2026clickenhance,zhang2026pinpoint3d,han2024clickformer,zhou2024pointsam,liu2025npiseg3d,simonelli2025easy3d,cong2026e2i3d,kang2026clickseg3d,zhang2026negroi}.

Despite recent progress, existing methods are still computationally redundant, as they process the entire scene after every interaction, even though the previous segmentation result and the latest click already provide strong spatial cues about the target object, which typically occupies only a small portion of the scene. 

In this paper, we first introduce \textbf{FastBox3D}, a dynamic localization strategy for efficient interactive 3D segmentation, to address this limitation in efficiency. 
The base pipeline voxelizes and encodes the scene, followed by transformer-based interaction decoding. From the second click onward, it restricts the decoding to a local 3D region it constructs from the previous segmentation result and the latest user feedback. By avoiding repeated processing of scene features unrelated to the target object, FastBox3D substantially reduces computational cost while maintaining segmentation accuracy.

\begin{figure}[t]
    \centering
    
    \includegraphics[
        width=\columnwidth,
        trim=20mm 20mm 20mm 13mm,
        clip]{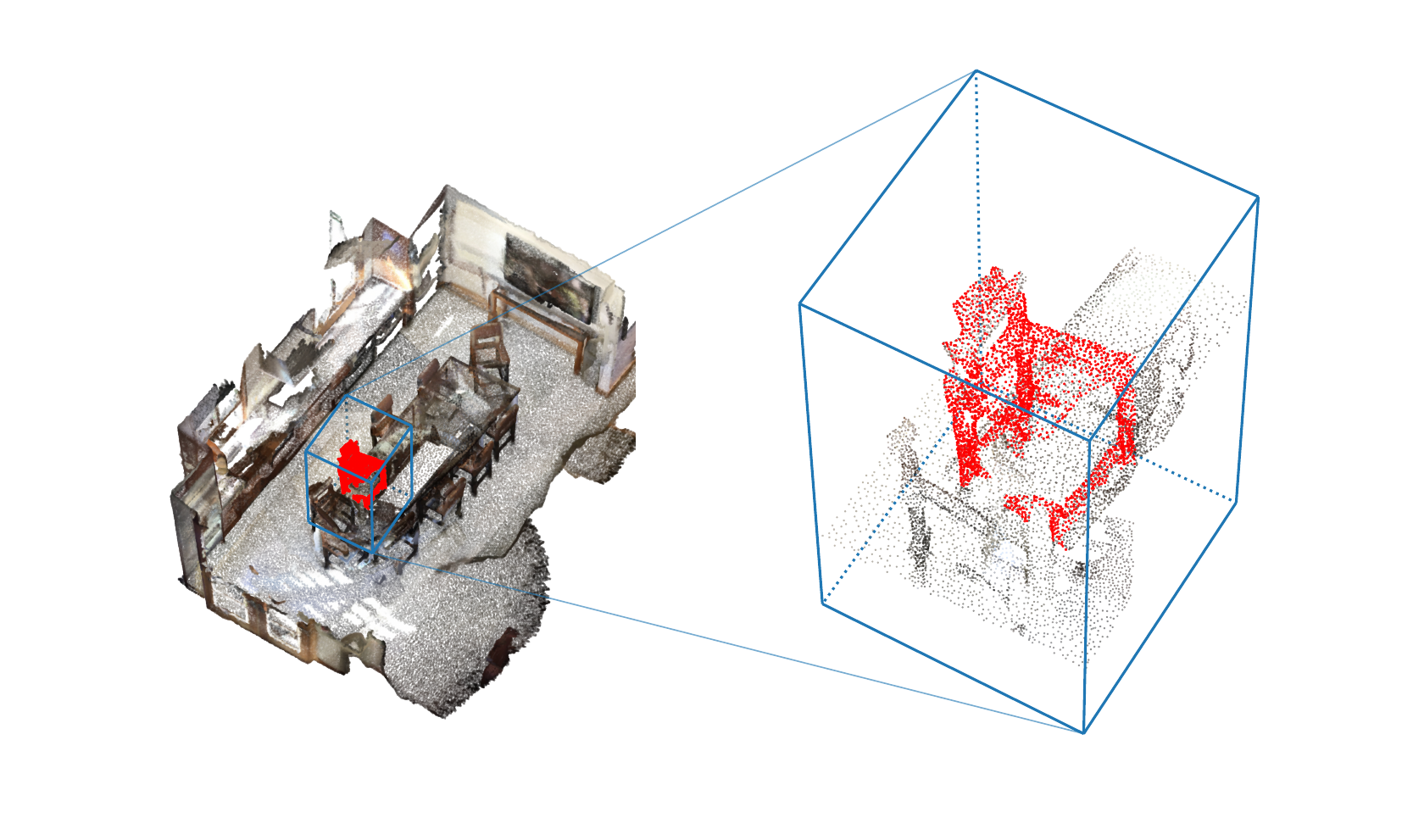}
    \caption{Conceptual illustration of spatially localized processing in HDBox3D. The blue box defines a compact region around the target object, and the right panel shows an enlarged view of the selected region. Points outside the selected region are excluded from subsequent processing.
}
    \label{fig:teaser}
\end{figure}

Taking advantage of the computational savings achieved by FastBox3D, we then propose \textbf{HDBox3D}, which reallocates a part of this saved computation to selectively increase the spatial resolution around the predicted target object. After the first prediction, HDBox3D re-voxelizes and re-encodes an expanded target region at a finer resolution, and it performs subsequent localized decoding using these high-resolution features. These features are reused across interactions and recomputed only when the required region extends beyond the current high-resolution area. This enables fine-grained segmentation while preserving computational efficiency.

In addition, we introduce \textbf{Voxel Size Alignment (VSA)} to address performance degradation caused by differences in point cloud sparsity between the training and inference domains. Our analysis of KITTI-360 reveals that using the same voxel resolution across domains can degrade segmentation performance when the target point cloud is substantially sparser than the training data. VSA adjusts the voxel size according to the relative point spacing between the training and target point clouds, aligning the voxel resolution with the density of the target domain. Despite its simplicity, VSA effectively mitigates performance degradation under substantial point-density mismatch.

Our contributions are summarized as follows:
\begin{itemize}
\item We introduce FastBox3D, which dynamically localizes interactive decoding using the previous mask and the latest click, substantially reducing computational cost.
\item We propose HDBox3D, which extends FastBox3D with selective high-resolution encoding and decoding around the target object, improving segmentation accuracy while maintaining efficiency.
\item We introduce Voxel Size Alignment (VSA), which adapts the voxel size to differences in point cloud sparsity between the training and target data, improving segmentation accuracy when the target data are considerably sparser than the training data.
\end{itemize}

Our method achieves state-of-the-art or competitive performance across all four evaluated datasets: ScanNet~\cite{dai2017scannet}, S3DIS~\cite{armeni2016s3dis}, KITTI-360~\cite{liao2022kitti360}, and ScanNet++~\cite{yeshwanth2023scannetpp}.

\section{Related work}
\label{sec:related_work}

\paragraph{Interactive 3D instance segmentation.}
This task iteratively refines an object mask based on user feedback indicating which points should be included in (positive) or excluded from (negative) the mask. It was introduced by InterObject3D~\cite{kontogianni2023interobject3d}, which enables users to provide positive and negative mouse clicks directly on a 3D point cloud. Subsequent methods have improved segmentation accuracy and computational efficiency~\cite{yue2024agile3d,zhang2026pinpoint3d,han2024clickformer,zhou2024pointsam,liu2025npiseg3d,simonelli2025easy3d,cong2026e2i3d,kang2026clickseg3d,zhang2026negroi}.

AGILE3D~\cite{yue2024agile3d} voxelizes the input point cloud and employs a sparse U-Net encoder with a Transformer-based decoder. It uses explicit click fusion, combining the masks predicted from individual clicks using a max operation. Point-SAM~\cite{zhou2024pointsam}, in contrast, groups the input points into patch tokens using a Voronoi tokenizer and processes them with a ViT encoder. It follows SAM~\cite{kirillov2023sam} in using a decoder based on a two-way Transformer and implicit click fusion. Easy3D~\cite{simonelli2025easy3d} builds upon AGILE3D and Point-SAM by combining the voxel-based sparse U-Net encoder of AGILE3D with the SAM-style decoder and implicit click fusion used in Point-SAM. ClickFormer~\cite{han2024clickformer} addresses large-scale variations among target regions, while NPISeg3D~\cite{liu2025npiseg3d} introduces probabilistic modeling for few-click segmentation and uncertainty estimation. SNAP~\cite{gupta2026snap} further improves cross-domain generalization across diverse point-cloud domains. More recently, E2I3D~\cite{cong2026e2i3d} improves efficiency through heterogeneous pruning and uses hierarchical click-aware attention to integrate fine geometric details with global context, while NegROI~\cite{zhang2026negroi} performs click-centric multi-resolution processing within a local ROI and introduces scene-conditioned negative prompts to suppress hard false positives.

\paragraph{Spatially restricted interactive segmentation.}
Restricting the processing region has been explored in 2D interactive segmentation to reduce unnecessary computation and focus processing on regions relevant to the target object. f-BRS~\cite{sofiiuk2020fbrs} employs a Zoom-In strategy that focuses subsequent inference on the predicted object region. FocusCut~\cite{lin2022focuscut} performs local refinement on adaptively selected regions around corrective clicks, while FocalClick~\cite{chen2022focalclick} introduces a Target Crop for object segmentation and a Focus Crop for local correction. These approaches demonstrate that restricting the processing region can substantially reduce redundant computation during interaction. 

In this paper, we demonstrate that such spatial restriction is particularly important for 3D point clouds, where large portions of a scene can be spatially distant from and irrelevant to the target object.

\paragraph{Localized high-resolution processing.}
Beyond spatial restriction, selective high-resolution processing has been explored to recover fine-grained details without processing the entire input at high resolution. In 2D image segmentation, PointRend~\cite{kirillov2020pointrend} refines uncertain locations in a coarse-to-fine manner, while CascadePSP~\cite{cheng2020cascadepsp} combines global prediction with local high-resolution refinement. NegROI~\cite{zhang2026negroi} applies a related strategy to interactive 3D segmentation by performing full-scene decoding followed by fine-resolution processing of a click-centric ROI.

In contrast, our HDBox3D performs full-scene decoding only for the first click and restricts decoding to a localized region afterwards. Moreover, its high-resolution features are reused across interactions and recomputed only when the required region extends beyond the cached area, avoiding repeated global decoding and fine-resolution encoding.

\paragraph{Density-aware point cloud processing.}
Several papers have addressed domain gaps caused by variations in LiDAR sampling density through density manipulation~\cite{wei2022lidardistillation,li2024densityresampling,kim2024multidensity}, while density-aware methods explicitly incorporate density information into voxel representations~\cite{hu2022pdv,kim2024densityaware}.
Cylinder3D~\cite{zhu2021cylinder3d} addresses non-uniform LiDAR point density through cylindrical spatial partitioning, and adaptive voxelization has also been explored to improve point sampling and scene representation~\cite{yang2025avsnet}.

Unlike these approaches, our Voxel Size Alignment adjusts only the inference-time voxel size according to the relative point spacing between the training and target data, without additional training or parameter updates.

\def\VE{\mathbf{V}_{\hspace{-2pt}E}}
\def\VEH{\widehat{\mathbf{V}}_{\hspace{-2pt}E}}

\section{Method}
\label{sec:method}

In this section, we first present the preliminaries of our method (\cref{sec:preliminaries}), then introduce FastBox3D (\cref{sec:box3d}). We next detail HDBox3D (\cref{sec:hdbox3d}), and finally introduce Voxel Size Alignment (\cref{sec:vsa}).

\begin{figure}[t]
    \centering
    \includegraphics[width=\columnwidth]{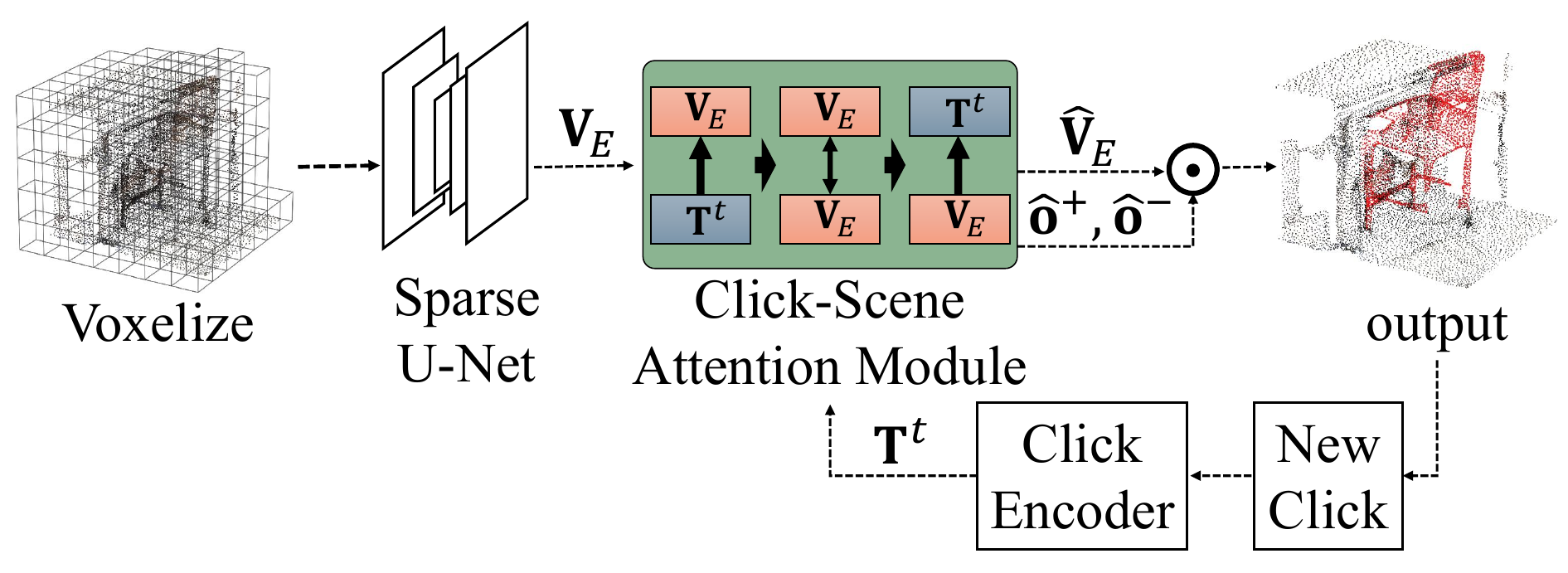}
    \caption{Overview of the base architecture. The input point cloud is first voxelized and encoded by a sparse U-Net to obtain scene embeddings $\VE$, which are computed once and reused throughout the interaction process. User clicks are encoded into decoder tokens $\mathbf{T}^t$ and jointly processed with the scene embeddings by a Click-Scene Attention Module. The updated voxel embeddings $\VEH$ and positive and negative output tokens $\widehat{\mathbf{o}}^{+}$ and $\widehat{\mathbf{o}}^{-}$ are used to predict the segmentation mask. Based on the current prediction, a new click is provided and the mask is iteratively refined.
}
    \label{fig:base_architecture}
\end{figure}

\subsection{Preliminaries}
\label{sec:preliminaries}
Our base architecture is illustrated in \cref{fig:base_architecture}. It follows Easy3D~\cite{simonelli2025easy3d}.

Let a 3D scene be represented by a set
$\mathcal{S}=\{(\mathbf{x}_i^\mathrm{p},\mathbf{f}_i^\mathrm{p})\}_{i=1}^{N_\mathrm{p}}$ of $N_\mathrm{p}$ points, where
$\mathbf{x}_i^\mathrm{p}\in\mathbb{R}^{3}$ and $\mathbf{f}_i^\mathrm{p}\in\mathbb{R}^{3}$ are the 3D coordinates and RGB values of the $i$-th point, respectively.
At interaction step $t$, the user provides a set
$\mathcal{C}^t=\{(\mathbf{x}_j^\mathrm{c},l_j)\}_{j=1}^{N_\mathrm{c}^t}$ of $N_\mathrm{c}^t$ clicks, where
$\mathbf{x}_j^\mathrm{c}\in\mathbb{R}^{3}$ is the 3D position of the $j$-th click and $l_j$ is a binary label indicating whether the clicked point should be in the mask (positive), or out of it (negative).
Given the scene $\mathcal{S}$ and the accumulated clicks $\mathcal{C}^t$, the objective is to predict a binary point mask
$\mathbf{M}_\mathrm{p}^t\in\{0,1\}^{N_\mathrm{p}}$ that represents the segmentation.

To efficiently process the 3D scene, the input point cloud is first voxelized using a fixed voxel size.
This produces a set $\mathcal{V}=\{(\mathbf{x}_i^\mathrm{v},\mathbf{f}_i^\mathrm{v})\}_{i=1}^{N_\mathrm{v}}$ of $N_\mathrm{v}$ occupied voxels, where
$\mathbf{x}_i^\mathrm{v}\in\mathbb{R}^{3}$ and $\mathbf{f}_i^\mathrm{v}\in\mathbb{R}^{3}$ represent the normalized 3D coordinates and RGB values of the $i$-th voxel, respectively.
The click positions are mapped to the same voxelized scene coordinates.
We denote the voxel-level prediction at interaction step $t$ by
$\mathbf{M}_\mathrm{v}^t\in\{0,1\}^{N_\mathrm{v}}$.

We then pass the voxelized scene to a sparse U-Net encoder~\cite{choy2019minkowski} $E$ to obtain the corresponding voxel embeddings,
$\VE = E(\mathcal{V})\in\mathbb{R}^{N_\mathrm{v}\times d}$, where $d$ denotes the embedding dimension.
Since the scene encoding is independent of user interactions, $\VE$ is computed only once and reused throughout the subsequent interaction steps.

Meanwhile, the click encoder transforms the accumulated clicks $\mathcal{C}^t$ into
click embeddings $\mathbf{C}_E^t\in\mathbb{R}^{N_\mathrm{c}^t\times d}$.
Following Easy3D~\cite{simonelli2025easy3d}, each click is represented by combining a positional encoding of its 3D location with a learned encoding of its positive or negative label.
Specifically, the embedding of the $j$-th click is given by $\mathbf{e}_j=\phi(\mathbf{x}_j^\mathrm{c})+\mathbf{e}_{l_j}$, where $\phi(\mathbf{x}_j^\mathrm{c})$ is the positional encoding of the click location and
$\mathbf{e}_{l_j}$ denotes the learned label encoding.
The resulting click embeddings are combined with learned positive and negative output tokens
$\mathbf{o}^{+}$ and $\mathbf{o}^{-}$ to form the decoder tokens $\mathbf{T}^t$.

The scene and interaction representations are then jointly updated by the Click-Scene Attention Module decoder $D$.
This module is based on a Two-Way Transformer~\cite{kirillov2023sam} that combines self-attention and cross-attention operations, enabling bidirectional interaction between the voxel embeddings and the decoder tokens.
The result is $(\VEH,\widehat{\mathbf{T}}^t)=D(\VE,\mathbf{T}^t)$.
Let $\widehat{\mathbf{v}}_i$ denote the updated embedding of the $i$-th voxel, and
$\widehat{\mathbf{o}}^{+}$ and $\widehat{\mathbf{o}}^{-}$ denote the updated positive and negative output tokens contained in $\widehat{\mathbf{T}}^t$.

Finally, foreground and background scores for each voxel are obtained by the dot products between its updated embedding and the corresponding output tokens,
\begin{equation}
    s_i^{+}
    =
    \left\langle
    \widehat{\mathbf{v}}_i,
    \widehat{\mathbf{o}}^{+}
    \right\rangle,
    \qquad
    s_i^{-}
    =
    \left\langle
    \widehat{\mathbf{v}}_i,
    \widehat{\mathbf{o}}^{-}
    \right\rangle.
\end{equation}
The voxel mask is obtained as
\begin{equation}
    m_{\mathrm{v},i}^t
    =
    \begin{cases}
        1, & s_i^{+} > s_i^{-},\\
        0, & \text{otherwise},
    \end{cases}
\end{equation}
where $m_{\mathrm{v},i}^t$ denotes the $i$-th element of $\mathbf{M}_\mathrm{v}^t$.
The voxel mask is then mapped back to the original points using the voxelization correspondence, yielding the final point mask $\mathbf{M}_\mathrm{p}^t$.

\subsection{FastBox3D}
\label{sec:box3d}

\begin{figure}[t]
    \centering
    \includegraphics[
        width=\columnwidth,
        trim=10mm 40mm 10mm 26mm,
        clip
    ]{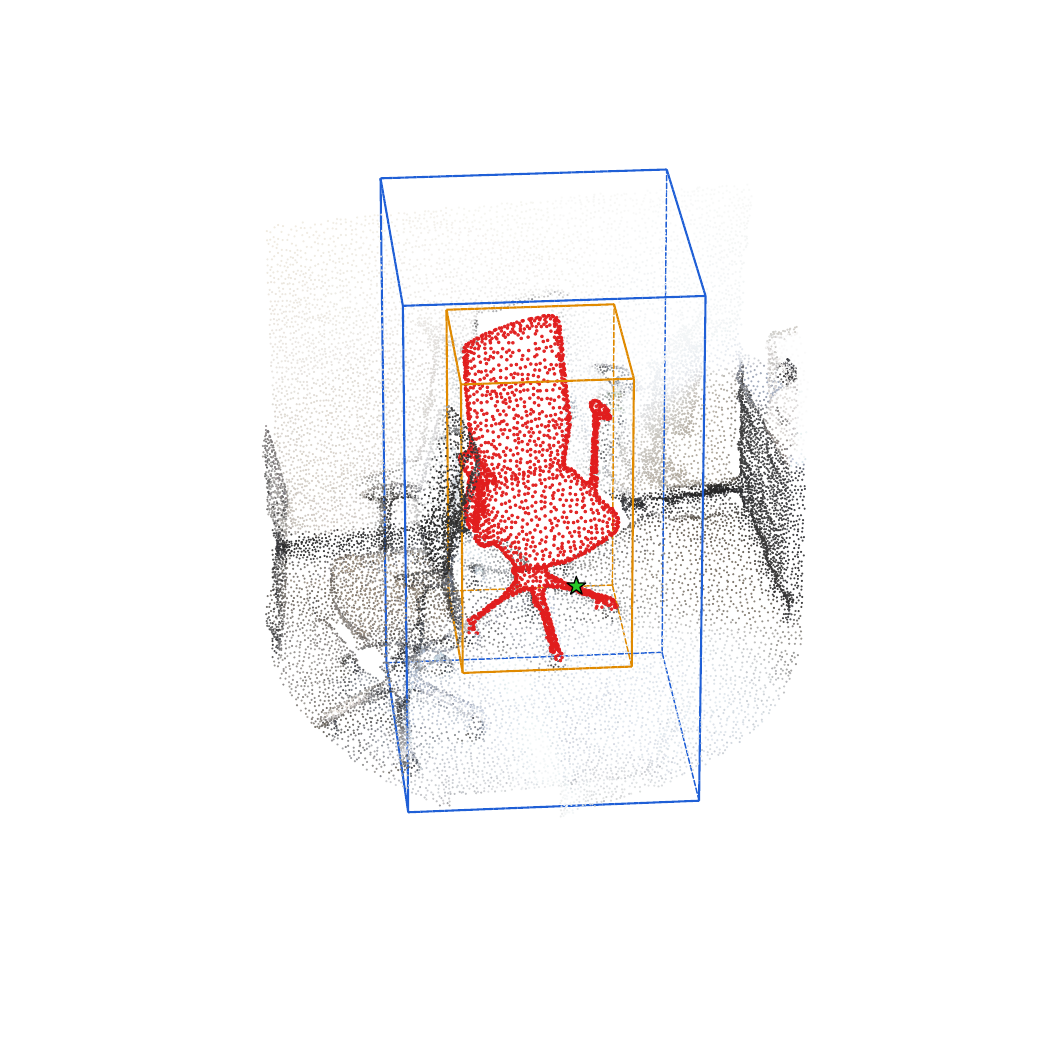}
    \caption{
Local region construction in FastBox3D.
The yellow box represents the minimal axis-aligned bounding box enclosing the previous foreground prediction and the latest click (green star).
The blue box represents the expanded box obtained by enlarging the minimal box while keeping its center fixed.
}
    \label{fig:box3d}
\end{figure}

\begin{figure*}[t]
    \centering
    \includegraphics[width=\textwidth]{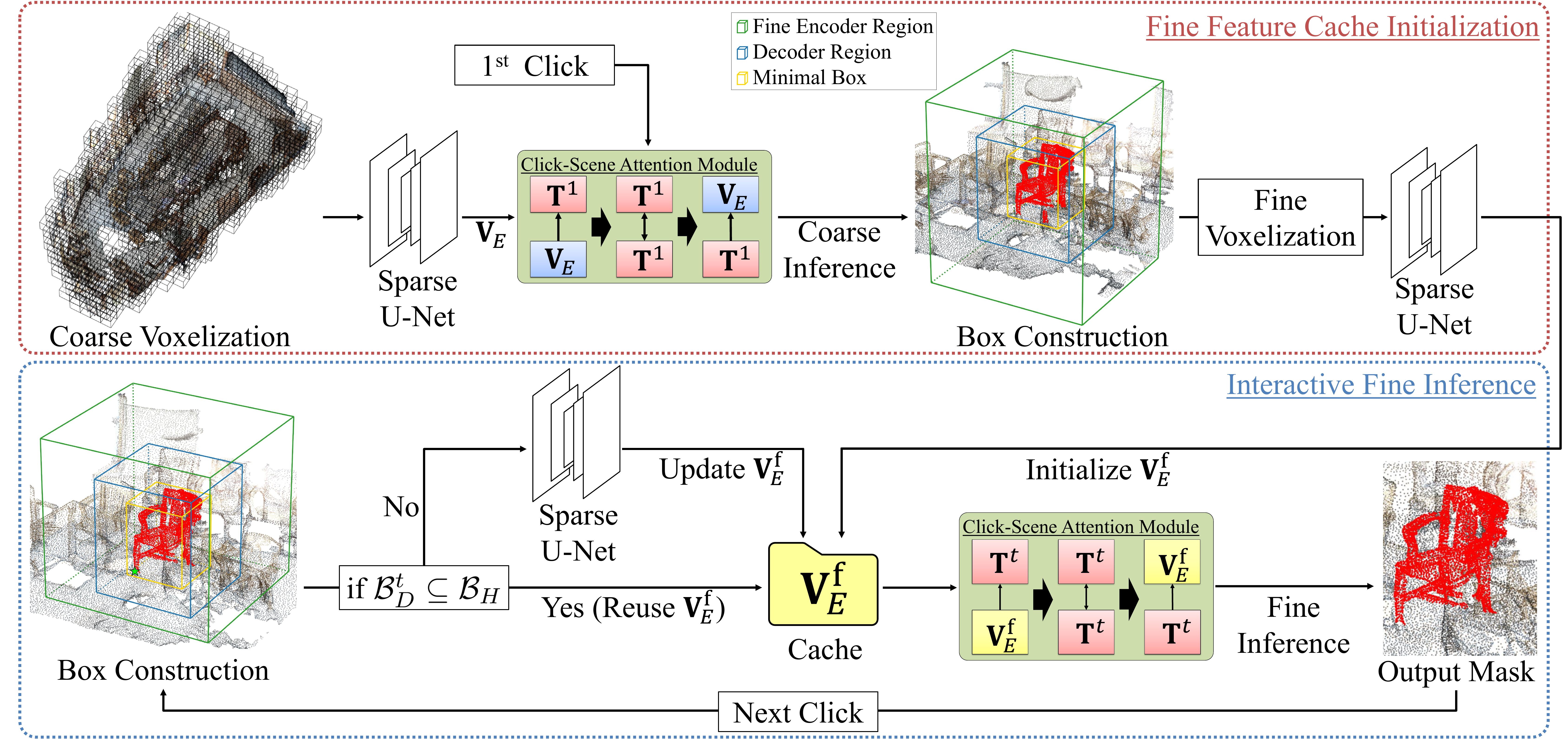}
    \caption{Overview of HDBox3D. In the Fine Feature Cache Initialization stage, the scene is coarsely voxelized and encoded by a sparse U-Net to obtain voxel embeddings $\VE$, followed by an initial coarse prediction using the Click-Scene Attention Module. This module is based on a Two-Way Transformer~\cite{kirillov2023sam} that combines self-attention and cross-attention operations. From this prediction, the minimal box (yellow) is expanded by $r_\mathrm{D}$ and $r_\mathrm{H}$ ($r_\mathrm{H}>r_\mathrm{D}$) to form the decoder region (blue) and fine encoding region (green), respectively. The fine encoding region is re-voxelized and encoded to obtain cached fine embeddings $\VE^\mathrm{f}$. In the Interactive Fine Inference stage, the first fine mask is predicted using the cached embeddings within the decoder region. From the second interaction onward, the decoder region is updated using the previous mask and the new click. The cache is reused while the decoder region remains within the cached fine encoding region; otherwise, the fine encoding region and cache are updated.
    }
    \label{fig:hdbox3d_overview}
\end{figure*}

In the base architecture, the decoder processes the entire scene at every interaction step, which is computationally redundant since subsequent corrections are typically limited to a local region around the current prediction and the newly added click.
Based on this observation, FastBox3D restricts the decoder computation to a dynamically constructed local region.
The first interaction is performed on the entire scene, as no previous prediction is available.
From the second interaction onward, FastBox3D constructs a local region using the previous voxel mask and the newly added click (\cref{fig:box3d}).

Let $\mathbf{x}_i^\mathrm{v}\in\mathbb{R}^{3}$ denote the spatial coordinate of the $i$-th occupied voxel, and let $\mathbf{x}_t^\mathrm{c}\in\mathbb{R}^{3}$ denote the position of the newly added click at interaction step $t$.
For $t\geq2$, we define the coordinate set used to construct the local region as
\begin{equation}
    \mathcal{X}^{t}
    =
    \left\{
        \mathbf{x}_i^\mathrm{v}
        \mid
        m_{\mathrm{v},i}^{t-1}=1
    \right\}
    \cup
    \left\{
        \mathbf{x}_t^\mathrm{c}
    \right\},
\end{equation}
which contains the coordinates of the previous foreground prediction and the newly added click.
We then compute the minimum and maximum coordinates of $\mathcal{X}^{t}$ along each spatial axis as $\mathbf{x}_{\min}^{t} = \min_{\mathbf{x}\in\mathcal{X}^{t}}\mathbf{x}$ and $\mathbf{x}_{\max}^{t} = \max_{\mathbf{x}\in\mathcal{X}^{t}}\mathbf{x}$.
These coordinates define the minimal axis-aligned bounding box $\mathcal{B}_{\min}^{t}$ (yellow box in \cref{fig:box3d}).
Including the newly added click allows this box to extend beyond the previous foreground prediction when the click lies outside of it.

Using this tightly fitted box directly may exclude useful spatial context around the current prediction.
Therefore, we enlarge $\mathcal{B}_{\min}^{t}$ by a fixed expansion ratio $r_\mathrm{B}$.
Its center $\mathbf{c}^{t}$ and side lengths $\mathbf{s}^{t}$ are given by
\begin{equation}
    \mathbf{c}^{t}
    =
    \frac{\mathbf{x}_{\min}^{t}+\mathbf{x}_{\max}^{t}}{2},
    \qquad
    \mathbf{s}^{t}
    =
    \mathbf{x}_{\max}^{t}-\mathbf{x}_{\min}^{t}.
\end{equation}
Keeping the center fixed, we expand each side by $r_\mathrm{B}$ and obtain the lower and upper coordinates of the expanded box as
\begin{equation}
    \widetilde{\mathbf{x}}_{\min}^{t}
    =
    \mathbf{c}^{t}
    -
    \frac{r_\mathrm{B}\mathbf{s}^{t}}{2},
    \qquad
    \widetilde{\mathbf{x}}_{\max}^{t}
    =
    \mathbf{c}^{t}
    +
    \frac{r_\mathrm{B}\mathbf{s}^{t}}{2}.
\end{equation}
The region bounded by these coordinates is denoted by $\mathcal{B}^{t}$ (blue box in \cref{fig:box3d}) and is used for subsequent decoder inference.

We then select the scene embeddings whose voxel coordinates lie inside $\mathcal{B}^{t}$.
The voxel index set is
\begin{equation}
    \mathcal{I}^{t}
    =
    \left\{
        i
        \mid
        \widetilde{\mathbf{x}}_{\min}^{t}
        \leq
        \mathbf{x}_i^\mathrm{v}
        \leq
        \widetilde{\mathbf{x}}_{\max}^{t}
    \right\},
\end{equation}
where the inequalities denote the conjunction of those evaluated along each spatial axis.
The local scene embeddings are then formed by collecting the corresponding voxel embeddings, $\mathbf{V}_\mathrm{B}^{t} = [\mathbf{v}_i]_{i\in\mathcal{I}^{t}} \in \mathbb{R}^{|\mathcal{I}^{t}|\times d}$, where 
$\mathbf{v}_i\in\mathbb{R}^{d}$ denotes the embedding of the $i$-th voxel in $\VE$.
Accordingly, the decoder operates only on the embeddings within the expanded box, producing $(\widehat{\mathbf{V}}_\mathrm{B}^{t},\widehat{\mathbf{T}}^{t})=D(\mathbf{V}_\mathrm{B}^{t},\mathbf{T}^{t})$.

Finally, the decoder updates the prediction only for voxels inside the expanded box.
Let $\widetilde{m}_{\mathrm{v},i}^{t}$ denote the resulting prediction for voxel $i\in\mathcal{I}^{t}$.
The complete voxel mask is updated as
\begin{equation}
    m_{\mathrm{v},i}^{t}
    =
    \begin{cases}
        \widetilde{m}_{\mathrm{v},i}^{t},
        & i\in\mathcal{I}^{t},\\
        m_{\mathrm{v},i}^{t-1},
        & i\notin\mathcal{I}^{t}.
    \end{cases}
\end{equation}
Thus, the prediction outside the expanded box is preserved from the previous interaction, while the prediction inside the box is updated using the accumulated clicks.
As in the base architecture, the resulting voxel mask $\mathbf{M}_\mathrm{v}^t$ is mapped back to the original points to obtain $\mathbf{M}_\mathrm{p}^t$.

\subsection{HDBox3D}
\label{sec:hdbox3d}
FastBox3D substantially reduces decoder computation by restricting subsequent interactions to a local region. HDBox3D takes advantage of the computational savings achieved by FastBox3D to process the target region at a higher spatial resolution. Instead of processing the entire scene at a finer resolution, we selectively re-encode the localized region using finer voxels. In this way, the reduced computational cost of local decoding enables higher-resolution feature extraction while maintaining efficient interaction.

HDBox3D begins with the same full-scene inference as in the base architecture. Let $h_\mathrm{c}$ denote the edge length of each voxel used for this initial inference. At the first interaction, the complete scene is voxelized with $h_\mathrm{c}$ and encoded by the sparse U-Net. Then, it is processed by the Click-Scene Attention Module to obtain an initial mask $\mathbf{M}_\mathrm{v}^1$. This initial prediction is then used to localize the target object and define the region for finer feature extraction.

Specifically, we construct the minimum axis-aligned box that contains the predicted foreground and enlarge it using an expansion ratio $r_\mathrm{H}$, following the same box expansion procedure as in FastBox3D.   We denote the resulting region by $\mathcal{B}_H$. The original points inside $\mathcal{B}_H$ are extracted directly from the input scene and voxelized again using a finer voxel edge length $h_\mathrm{f}$, where $h_\mathrm{f}<h_\mathrm{c}$. The resulting voxel set $\mathcal{V}_H$ is passed through the same sparse U-Net encoder to obtain finer scene embeddings $\VE^\mathrm{f} = E(\mathcal{V}_H)$. Both the fine encoding region $\mathcal{B}_H$ and its corresponding fine embeddings $\VE^\mathrm{f}$ are cached for reuse in subsequent interactions.

The decoding region is smaller than the fine encoding region. Starting from the same minimum axis-aligned box of the predicted foreground, we construct another expanded box using a decoder expansion ratio $r_\mathrm{D}$, where $r_\mathrm{D}<r_\mathrm{H}$, and denote it by $\mathcal{B}_D^1$. Only the embeddings in $\VE^\mathrm{f}$ whose coordinates lie inside $\mathcal{B}_D^1$ are passed to the Click-Scene Attention Module. The resulting prediction replaces the initial coarse prediction and becomes the output of the first interaction. Using a larger region for fine encoding allows the finer embeddings to be reused as the decoder region changes over subsequent interactions, while restricting decoding to the smaller $\mathcal{B}_D^1$ reduces the computational cost of the Click-Scene Attention Module.

For subsequent interactions, HDBox3D follows the localization strategy of FastBox3D using the finer voxel representation. At interaction step $t\geq2$, the previous foreground prediction and the newly added click are used to construct a decoder region $\mathcal{B}_D^t$ with expansion ratio $r_\mathrm{D}$. If $\mathcal{B}_D^t$ is fully contained in the cached fine encoding region $\mathcal{B}_H$, the stored embeddings $\VE^\mathrm{f}$ are reused and only the embeddings inside $\mathcal{B}_D^t$ are selected for decoding. Consequently, full-scene decoding is avoided, while the stored fine embeddings can be reused across multiple clicks.

The decoder region $\mathcal{B}_D^t$ may extend beyond the cached fine encoding region $\mathcal{B}_H$. In this case, we construct a new fine encoding region $\mathcal{B}_H^t$ from the current foreground prediction and the newly added click, using the larger expansion ratio $r_\mathrm{H}$. The points inside $\mathcal{B}_H^t$ are voxelized with $h_\mathrm{f}$ and encoded again to refresh $\VE^\mathrm{f}$. The update can therefore be summarized as
\begin{equation}
    \VE^\mathrm{f}
    =
    \begin{cases}
        \VE^\mathrm{f},
        & \mathcal{B}_D^{t}\subseteq\mathcal{B}_H,\\
        E(\mathcal{V}_H^{t}),
        & \text{otherwise}.
    \end{cases}
\end{equation}
Here, $\mathcal{V}_H^{t}$ denotes the finer voxel representation constructed within the newly updated encoding region. After recomputation, the newly constructed region $\mathcal{B}_H^t$ becomes the cached fine encoding region, \ie, $\mathcal{B}_H \leftarrow \mathcal{B}_H^t$. This conditional update allows the finer scene features to persist across multiple clicks and invokes the sparse U-Net again only when the cached encoded region is no longer sufficient.

Overall, HDBox3D uses the computation saved by FastBox3D to extract higher-resolution features around the target object. By restricting fine encoding to a local region and reusing the resulting embeddings across interactions, it avoids full-scene high-resolution processing and repeated sparse U-Net execution. HDBox3D also retains FastBox3D's local decoding strategy. This improves target representation while preserving computational efficiency.

\subsection{Voxel size alignment}
\label{sec:vsa}

\begin{figure}[t]
    \centering
    \includegraphics[width=\columnwidth]{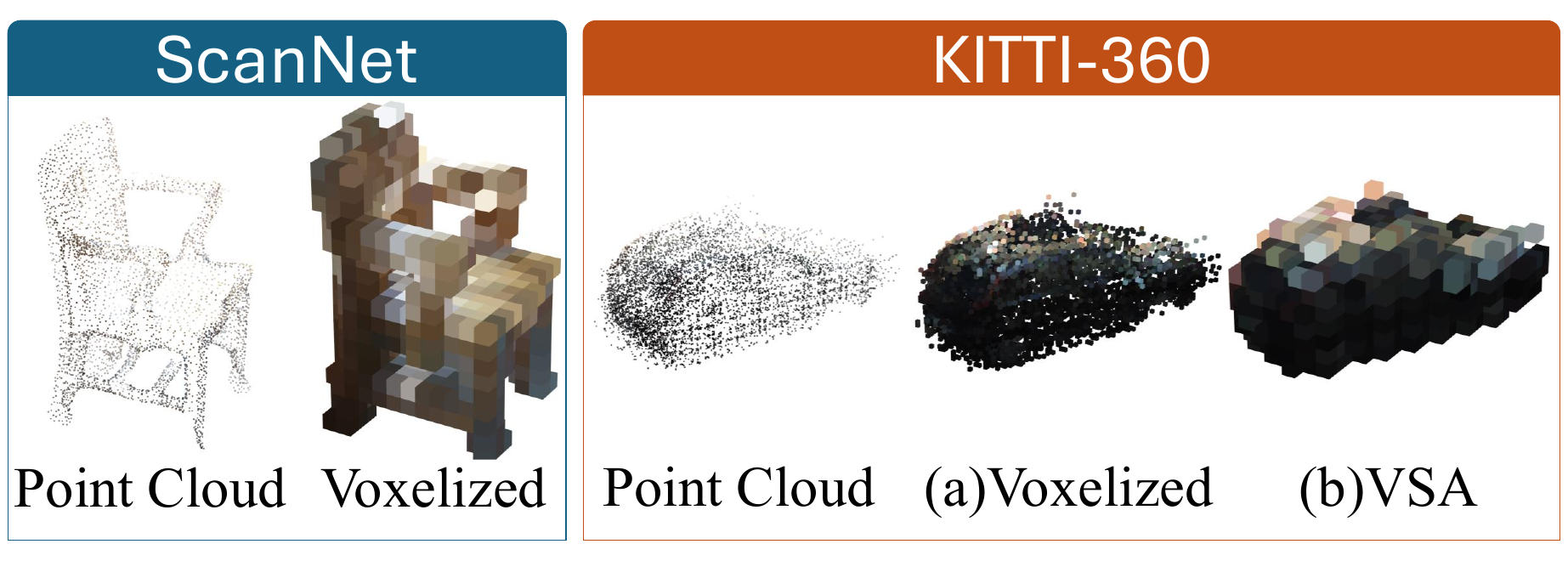}
    \caption{Comparison of voxelization across datasets with different point densities. Applying the ScanNet voxel size directly to the sparser KITTI-360 point cloud results in a sparse voxel representation (a). VSA increases the voxel size according to the relative point spacing, producing a denser voxel representation whose sparsity more closely matches that of ScanNet (b).
}
    \label{fig:vsa}
\end{figure}

Point clouds from different datasets can have different point densities.
When a model trained on dense point clouds is applied to a sparser cross-domain target with a fixed voxel size, the resulting voxel representation can become considerably sparser than during training (\cref{fig:vsa}), degrading segmentation performance.
To mitigate this density mismatch, we introduce Voxel Size Alignment (VSA), which dynamically adjusts the voxel size based on the point spacing ratio between the training data and each target scene.

We first compute the average nearest-neighbor distance across all points in the training dataset, denoted by $d_{\mathrm{train}}$.
During cross-domain inference on a target scene $S$, we similarly compute the average nearest-neighbor distance $d_{\mathrm{target}}(S)$ and define the spacing ratio as $r(S) = d_{\mathrm{target}}(S) / d_{\mathrm{train}}$.

When the target scene is sparser than the training data ($r(S) > 1$), VSA scales the voxel size by $r(S)$, while keeping it unchanged for denser scenes ($r(S) \le 1$).

\def\na{-}
\newcommand{\bestcell}[1]{\cellcolor{red!15}\textbf{#1}}
\newcommand{\secondcell}[1]{\cellcolor{yellow!15}#1}
\section{Experiments}
\label{sec:experiments}

In this section, we first describe the datasets (\cref{sec:datasets}) and evaluation protocol (\cref{sec:eval_protocol}), followed by the implementation details (\cref{sec:implementation}). We then present the quantitative (\cref{sec:quantitative}) and qualitative results (\cref{sec:qualitative}), and finally report ablation studies (\cref{sec:ablation}).

\begin{table*}
    \captionsetup{
        width=0.86\textwidth
    }
    \caption{
        Quantitative results on single-object interactive segmentation.
        All models are trained exclusively on ScanNet.
        $^{*}$ indicates methods using VSA,
        $^{\dagger}$ indicates results from the original papers,
        and ``\na'' denotes unavailable results.
        Best and second-best IoU values are highlighted in \colorbox{red!15}{\textbf{red}} and \colorbox{yellow!15}{yellow}, respectively.
    }
    \label{tab:main_results}
    \centering

    \resizebox{1.9\columnwidth}{!}{%
    \begin{tabular}{l|l|cccccc|cc|cc|cc}
        \toprule
        \multirow{2}{*}{\textbf{Test Dataset}} 
        & \multirow{2}{*}{\textbf{Method}} 
        & \multicolumn{6}{c|}{\textbf{IoU}} 
        & \multicolumn{2}{c|}{\textbf{GFLOPs}} 
        & \multicolumn{2}{c|}{\textbf{Latency (ms)}} 
        & \multicolumn{2}{c}{\textbf{Peak Mem (MiB)}} \\
        & & \textbf{@1} & \textbf{@2} & \textbf{@3} & \textbf{@5} & \textbf{@7} & \textbf{@10} 
        & \textbf{@1 $\downarrow$} 
        & \textbf{@2--10 $\downarrow$} 
        & \textbf{@1 $\downarrow$} 
        & \textbf{@2--10 $\downarrow$} 
        & \textbf{Avg. $\downarrow$} & \textbf{Max $\downarrow$} \\
        \midrule

        \multirow{6}{*}{ScanNet} 
        & AGILE3D~\cite{yue2024agile3d}                 & 63.0          & 71.1          & 75.7          & 80.1          & \secondcell{82.1} & \secondcell{83.7} & \na    & \na    & \na   & \na   & \na     & \na      \\
        & E$^2$I3D$^{\dagger}$~\cite{cong2026e2i3d}     & \na           & \na           & \na           & 79.9          & \na           & 83.6          & \na    & \na    & \na   & \na   & \na     & \na      \\
        & NegROI$^{\dagger}$~\cite{zhang2026negroi}     & \bestcell{72.1} & \secondcell{78.1} & \secondcell{80.9} & \secondcell{82.3} & \na           & 83.6          & \na    & \na    & \na   & \na   & \na     & \na      \\
        & Easy3D~\cite{simonelli2025easy3d}             & 69.4          & 75.8          & 78.4          & 80.5          & 81.5          & 82.2          & 20.905 & 20.744 & 16.76 & 16.45 & 339.951 & 710.620  \\
        & Ours (FastBox3D)                              & 69.8          & 75.9          & 78.7          & 80.8          & 81.7          & 82.5          & 20.905 & 1.800  & 16.80 & 4.73  & 310.283 & 735.454  \\
        & Ours (HDBox3D)                                & \secondcell{71.0} & \bestcell{78.2} & \bestcell{81.8} & \bestcell{84.9} & \bestcell{86.3} & \bestcell{87.4} & 33.527 & 5.899  & 48.78 & 8.07  & 387.678 & 1710.711 \\
        \hline

        \multirow{7}{*}{S3DIS} 
        & AGILE3D~\cite{yue2024agile3d}                 & 59.2          & 71.7          & 78.3          & 84.0          & 86.7          & \secondcell{88.6} & \na    & \na    & \na   & \na   & \na     & \na      \\
        & E$^2$I3D$^{\dagger}$~\cite{cong2026e2i3d}     & \na           & \na           & \na           & 84.2          & \na           & \secondcell{88.6} & \na    & \na    & \na   & \na   & \na     & \na      \\
        & Point-SAM$^{\dagger}$~\cite{zhou2024pointsam} & 38.8          & \na           & 67.1          & 72.2          & 78.9          & 80.6          & \na    & \na    & \na   & \na   & \na     & \na      \\
        & NegROI$^{\dagger}$~\cite{zhang2026negroi}     & 66.9          & 77.8          & 82.2          & 85.4          & \na           & 87.8          & \na    & \na    & \na   & \na   & \na     & \na      \\
        & Easy3D~\cite{simonelli2025easy3d}             & 68.9          & \secondcell{79.0} & 83.1          & 86.4          & 87.5          & 88.3          & 15.940 & 15.819 & 19.29 & 17.06 & 279.561 & 626.148  \\
        & Ours (FastBox3D)                              & \secondcell{69.2} & 78.9          & \secondcell{83.4} & \secondcell{86.6} & \secondcell{87.8} & \secondcell{88.6} & 15.940 & 1.927  & 20.00 & 5.07  & 255.867 & 677.275  \\
        & Ours (HDBox3D)                                & \bestcell{69.6} & \bestcell{79.6} & \bestcell{85.2} & \bestcell{89.7} & \bestcell{91.6} & \bestcell{93.0} & 31.381 & 7.631  & 51.87 & 9.36  & 370.468 & 2014.355 \\
        \hline

        \multirow{11}{*}{KITTI-360} 
        & AGILE3D~\cite{yue2024agile3d}                 & 32.0          & 41.1          & 44.6          & 47.7          & 50.5          & 54.6          & \na    & \na    & \na   & \na   & \na     & \na      \\
        & AGILE3D$^*$                                   & 44.1          & 56.0          & 61.3          & 66.1          & 69.2          & 73.7          & \na    & \na    & \na   & \na   & \na     & \na      \\
        & E$^2$I3D$^{\dagger}$~\cite{cong2026e2i3d}     & \na           & \na           & \na           & 49.6          & \na           & 52.2          & \na    & \na    & \na   & \na   & \na     & \na      \\
        & Point-SAM$^{\dagger}$~\cite{zhou2024pointsam} & 44.0          & \na           & 67.1          & 72.2          & 78.9          & 80.8          & \na    & \na    & \na   & \na   & \na     & \na      \\
        & NegROI$^{\dagger}$~\cite{zhang2026negroi}     & 47.5          & 62.6          & \secondcell{71.4} & \bestcell{80.1} & \na           & \bestcell{87.7} & \na    & \na    & \na   & \na   & \na     & \na      \\
        & Easy3D~\cite{simonelli2025easy3d}             & 40.0          & 51.8          & 61.5          & 72.7          & 78.8          & 82.9          & 20.235 & 20.079 & 16.08 & 15.74 & 329.384 & 1260.875 \\
        & Easy3D$^*$                                    & 48.4          & 63.2          & 69.8          & 76.2          & 79.3          & 81.5          & 3.528  & 3.510  & 5.36  & 5.11  & 130.498 & 427.727  \\
        & Ours (FastBox3D)                              & 45.5          & 53.1          & 61.6          & 73.7          & 79.7          & 84.2          & 20.235 & 4.708  & 16.02 & 5.77  & 309.111 & 1129.986 \\
        & Ours (FastBox3D$^*$)                          & \bestcell{54.0} & \bestcell{65.6} & \bestcell{71.7} & 77.6          & \secondcell{80.2} & 82.3          & 3.528  & 0.841  & 5.51  & 3.66  & 124.213 & 387.377  \\
        & Ours (HDBox3D)                                & 27.6          & 32.6          & 43.3          & 57.9          & 68.8          & 78.4          & 53.291 & 13.683 & 45.82 & 12.45 & 412.006 & 1343.369 \\
        & Ours (HDBox3D$^*$)                            & \secondcell{53.7} & \secondcell{63.5} & 70.4          & \secondcell{78.4} & \bestcell{82.2} & \secondcell{85.0} & 8.413  & 3.325  & 29.60 & 7.23  & 209.554 & 870.514  \\
        \hline

        \multirow{4}{*}{ScanNet++} 
        & AGILE3D~\cite{yue2024agile3d}                 & 49.8          & 58.4          & 63.8          & 70.0          & \secondcell{73.2} & \secondcell{75.8} & \na    & \na    & \na    & \na   & \na     & \na      \\
        & Easy3D~\cite{simonelli2025easy3d}             & \secondcell{57.6} & 63.6          & 67.0          & 70.4          & 72.3          & 73.8          & 56.166 & 55.714 & 71.37  & 46.69 & 789.956 & 2036.040 \\
        & Ours (FastBox3D)                              & \secondcell{57.6} & \secondcell{64.4} & \secondcell{67.8} & \secondcell{71.1} & 73.0          & 74.7          & 56.166 & 1.353  & 71.52  & 4.47  & 706.967 & 1810.487 \\
        & Ours (HDBox3D)                                & \bestcell{59.3} & \bestcell{67.3} & \bestcell{72.3} & \bestcell{77.2} & \bestcell{79.6} & \bestcell{81.5} & 72.357 & 5.072  & 104.71 & 7.45  & 825.973 & 3967.258      \\

        \bottomrule
    \end{tabular}}
\end{table*}

\subsection{Datasets}
\label{sec:datasets}
Following Easy3D~\cite{simonelli2025easy3d}, we evaluate our method on four 3D datasets: ScanNet~\cite{dai2017scannet}, S3DIS~\cite{armeni2016s3dis}, KITTI-360~\cite{liao2022kitti360}, and ScanNet++~\cite{yeshwanth2023scannetpp}.
ScanNet, S3DIS, and ScanNet++ contain indoor scenes, while KITTI-360 consists of outdoor urban scenes for autonomous driving.
For ScanNet++, we exclude objects containing fewer than eight voxels after voxelization with a voxel size of $0.05$\,m.
We train all models exclusively on ScanNet and evaluate them on all four datasets.

\subsection{Evaluation protocol}
\label{sec:eval_protocol}
Following previous work~\cite{yue2024agile3d,simonelli2025easy3d}, we adopt the single-object evaluation setting, where each evaluation targets one object instance in a 3D scene.
Under this setting, we evaluate segmentation accuracy using IoU@$i$, which denotes the mean Intersection over Union between the predicted and ground-truth instance masks after $i$ clicks.

\paragraph{Automatic click simulation.}
We use the same automatic click simulation policy as in previous work~\cite{yue2024agile3d,simonelli2025easy3d} during training and evaluation.
The first click is placed at the center of the target object, defined as the point farthest from any point outside the object.
Subsequent clicks are placed at the center of the largest error region and labeled positive for false negatives and negative for false positives.
This procedure is repeated for up to 10 clicks per object.

\paragraph{Computational efficiency.}
We evaluate computational efficiency in terms of GFLOPs, inference latency, and peak GPU memory.
GFLOPs are counted for the baseline Easy3D and our FastBox3D and HDBox3D, using analytical operation counting based on the actual sparse-convolution rulebooks and decoder token counts, where one multiply--accumulate is counted as one FLOP.
We exclude the initial full-scene sparse U-Net encoding from the reported costs, since it is identical across Easy3D, FastBox3D, and HDBox3D under the same voxel-size setting.
Thus, the reported costs for Easy3D and FastBox3D include only the per-click computation of the Click-Scene Attention Module, while those for HDBox3D additionally include fine-region U-Net computation when required.
We measure per-click inference latency on a single A100 GPU dedicated exclusively to evaluation.
For all methods, latency is measured from the input of each click to the generation of the corresponding point mask.
For peak GPU memory, we report the average per-object peak memory and the maximum peak memory across all objects in each dataset.
Efficiency results are not reported for E$^2$I3D~\cite{cong2026e2i3d} and NegROI~\cite{zhang2026negroi}, as their official code is not publicly available.

\subsection{Implementation details}
\label{sec:implementation}
Our training objective follows Easy3D~\cite{simonelli2025easy3d} and combines the Dice loss~\cite{deng2018crispboundaries} with cross-entropy using equal weights. For each simulated 10-click sequence, the mask predicted at every interaction contributes equally to the objective. We average the per-click losses over the sequence and update the model with a single backward pass.

For FastBox3D, we use AdamW~\cite{loshchilov2019adamw} with an initial learning rate of $1\times10^{-4}$.
The voxel edge length is set to $0.05$\,m, and the expansion ratio is set to $1.7$ based on the ablation study in \cref{sec:ablation}.

For HDBox3D, we initialize the model from the trained FastBox3D and fine-tune it with an initial learning rate of $1\times10^{-5}$.
The coarse and fine voxel edge lengths are set to $0.05$\,m and $0.025$\,m, respectively.
The decoder expansion ratio $r_\mathrm{D}$ is set to $1.7$, while the fine encoding expansion ratio $r_\mathrm{H}$ is set to $2.9$, obtained by further expanding the decoder region by a factor of $1.7$.
The coarse and fine stages share the weights of the sparse U-Net, while separate Click-Scene Attention Modules are used for the two stages.

For VSA, following the procedure described in \cref{sec:vsa}, we use the average nearest-neighbor distance of ScanNet ($0.013\,\mathrm{m}$) as $d_{\mathrm{train}}$. Among the cross-domain datasets, VSA is activated only for KITTI-360.

\begin{figure*}[t]
    \centering
    \includegraphics[width=\textwidth]{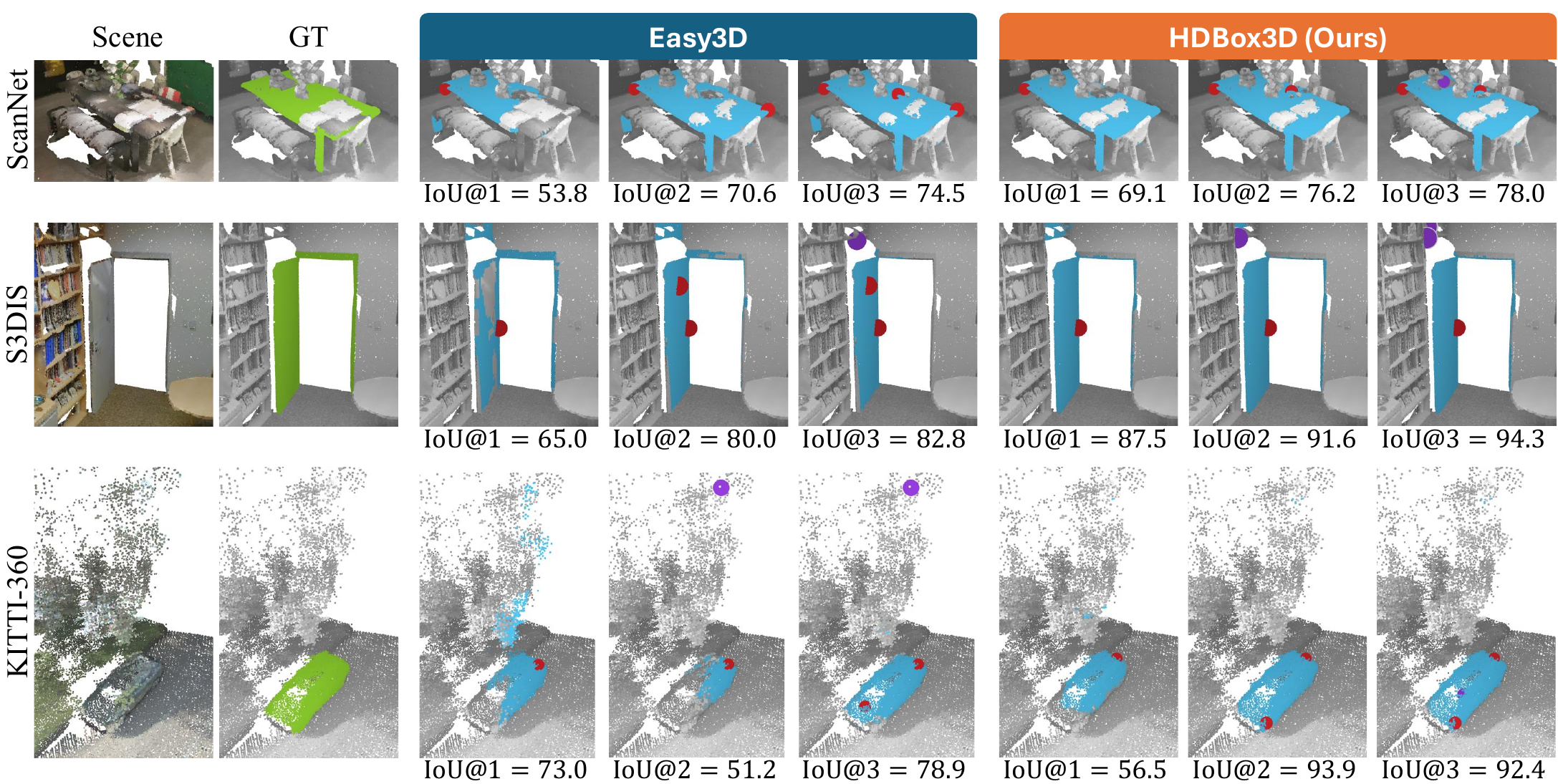}
    \caption{Qualitative comparison of Easy3D and HDBox3D on ScanNet, S3DIS and KITTI-360. For KITTI-360, HDBox3D is evaluated with Voxel Size Alignment (VSA). Red points indicate positive clicks, and purple points indicate negative clicks.
    }
    \label{fig:qualitative}
\end{figure*}

\subsection{Quantitative results}
\label{sec:quantitative}
\Cref{tab:main_results} compares our methods with existing interactive 3D segmentation methods.
We report GFLOPs and latency separately for the first click and as averages over clicks 2–10, since FastBox3D and HDBox3D use different computational pipelines for the first and subsequent clicks.

Compared with Easy3D, FastBox3D maintains segmentation accuracy, while HDBox3D further improves accuracy. On KITTI-360, HDBox3D without VSA suffers from a large performance drop due to the point sparsity discussed in \cref{sec:vsa}, but its accuracy substantially improves with VSA. Moreover, VSA improves performance for different methods across most click settings, indicating that its effectiveness is not specific to our architecture.

In terms of efficiency, FastBox3D substantially reduces FLOPs and latency from the second click onward and also lowers the average peak memory across all datasets. 
HDBox3D incurs additional computation, latency, and memory due to fine-resolution U-Net encoding, but localized decoding and fine feature caching substantially reduce its subsequent-click FLOPs. On ScanNet, S3DIS, and ScanNet++, these savings outweigh the initial overhead, resulting in lower cumulative FLOPs and latency over 10 clicks than Easy3D. On KITTI-360, VSA substantially reduces FLOPs, latency, and memory for all methods, thereby limiting the relative benefits of localized decoding and fine feature caching in HDBox3D; consequently, HDBox3D with VSA has a higher 10-click cumulative cost than Easy3D with VSA. 
Fine feature caching reduces the average number of fine U-Net executions to fewer than 1.5 per 10-click object sequence, compared with 10 executions when fine features are recomputed at every click.

\subsection{Qualitative results}
\label{sec:qualitative}
Qualitative results are shown in \cref{fig:qualitative}.
Compared with Easy3D, HDBox3D produces more accurate masks with fewer interactions.

\subsection{Ablation studies}
\label{sec:ablation}
\paragraph{Effect of expansion ratio.}
\begin{table}[t]
    \caption{Effect of the expansion ratio.}
    \label{tab:expansion_ratio}
    \centering
    \resizebox{\columnwidth}{!}{%
    \begin{tabular}{l|c|c|c|c|c}
        \toprule
        \textbf{Test Dataset} &
        \textbf{\begin{tabular}[c]{@{}c@{}}Expansion \\ Ratio\end{tabular}} &
        \textbf{IoU@3} &
        \textbf{IoU@5} &
        \textbf{IoU@10} &
        \textbf{GFLOPs $\downarrow$} \\
        \midrule
        \multirow{5}{*}{ScanNet}
            & 1.1 & 78.4          & 80.7          & \textbf{82.6} & \textbf{1.08} \\
            & 1.3 & 78.5          & 80.7          & 82.4          & 1.32          \\
            & 1.5 & 78.5          & 80.6          & 82.5          & 1.52          \\
            & 1.7 & \textbf{78.7} & \textbf{80.8} & 82.5          & 1.80          \\
            & 1.9 & 78.6          & 80.7          & 82.4          & 2.08          \\
        \bottomrule
    \end{tabular}}
\end{table}

We evaluate the effect of the expansion ratio on ScanNet, with decoder GFLOPs averaged over clicks 2--10. As shown in \cref{tab:expansion_ratio}, smaller expansion ratios reduce decoder computation, while segmentation accuracy remains relatively stable across different settings. A ratio of 1.7 achieves the best IoU@3 and IoU@5 at low computational cost, so we use it for all datasets without dataset-specific tuning. Results on the other datasets are provided in the supplementary material.

\paragraph{Effect of weight sharing.}
We evaluate the effect of weight sharing between the coarse and fine stages on ScanNet. The results on ScanNet are shown in \cref{tab:weight_sharing}, while the full results on all datasets are provided in the supplementary material. In the table, CSA Module denotes the Click-Scene Attention Module. Sharing both the U-Net and CSA Module reduces the parameter count from 27.77M to 13.89M while leaving segmentation accuracy nearly unchanged.
This indicates that the performance gain of HDBox3D is not simply explained by increased model capacity. We use shared U-Net weights with separate CSA Modules, which achieves the best accuracy with substantially fewer parameters than the fully separate model.

\begin{table}[t]
    \caption{Effect of weight sharing between coarse and fine stages.}
    \label{tab:weight_sharing}
    \centering
    \resizebox{\columnwidth}{!}{%
    \begin{tabular}{c|c|c|c|c|c}
        \toprule
        \textbf{U-Net} &
        \textbf{CSA Module} &
        \textbf{IoU@3} &
        \textbf{IoU@5} &
        \textbf{IoU@10} &
        \textbf{\#Params} \\
        \midrule
        Separate & Separate & 81.7 & 84.8 & 87.3 & 27.77M \\
        Shared   & Separate & \textbf{81.8} & \textbf{84.9} & \textbf{87.4} & 16.81M \\
        Shared   & Shared   & 81.6 & 84.7 & 87.3 & 13.89M \\
        \bottomrule
    \end{tabular}}
\end{table}

\section{Conclusion}
\label{sec:conclusion}
In this work, we presented FastBox3D and HDBox3D for efficient interactive 3D point cloud segmentation. FastBox3D restricts decoding to a localized region, reducing computation while maintaining accuracy. HDBox3D further improves segmentation by extracting and reusing high-resolution local features. We also introduced Voxel Size Alignment (VSA) to mitigate point-density differences across datasets. Experiments on four datasets demonstrate strong accuracy--efficiency trade-offs and improved cross-domain performance.

{
    \small
    \bibliographystyle{ieeenat_fullname}
    \bibliography{main}
}

\clearpage
\setcounter{page}{1}
\maketitlesupplementary
\appendix

This supplementary material provides additional implementation details (\cref{sec:sup_implementation}), additional ablation studies (\cref{sec:suppl_ablation}), further qualitative results (\cref{sec:suppl_qualitative}), and a discussion of the limitations (\cref{sec:suppl_limitations}).

\section{Additional implementation details}
\label{sec:sup_implementation}

\paragraph{Network architecture.}
We first normalize the scene coordinates using a fixed spatial scale of $40$\,m.
This normalization is used only to stabilize the coordinate representation and does not crop or otherwise limit the spatial extent of the input scene.
The normalized scene is then processed by a sparse U-Net consisting of five resolution levels with channel dimensions $[32, 64, 96, 128, 160]$, where each level contains two residual blocks.

The resulting scene features are projected to $256$-dimensional embeddings and processed by the Click-Scene Attention Module together with the click embeddings.
The module uses a two-layer Two-Way Transformer with eight attention heads per layer and an MLP hidden dimension of $1024$.
To encode spatial information, we use harmonic positional encoding with $12$ frequency bands.
Specifically, for each target object, the voxel and click coordinates are expressed relative to the first click and divided by the same fixed spatial scale before the harmonic encoding is applied.

\paragraph{Training details.}
We train FastBox3D for $1024$ epochs and subsequently fine-tune HDBox3D for $200$ epochs using DistributedDataParallel on four GPUs.
During each iteration, each GPU processes one scene, from which we randomly sample up to four target instances.
For every sampled instance, we simulate a sequence of $10$ clicks to train the model over the complete interaction process.

For optimization, we use AdamW with a weight decay of $0.05$ and a polynomial learning-rate schedule with a power of $0.9$.
We also apply both appearance and geometric augmentations during training.
For appearance augmentation, Gaussian noise with a standard deviation of $0.1$ is added to the RGB values.
For geometric augmentation, we first construct a linear transformation
$\mathbf{A}=\mathbf{I}+\boldsymbol{\epsilon}$,
where each entry of $\boldsymbol{\epsilon}$ is independently sampled from
$\mathcal{N}(0,0.1^2)$.
We then compose this transformation with a random reflection along the $x$-axis and a rotation around the $z$-axis, whose angle is sampled uniformly from $[0,2\pi)$.

Finally, for both FastBox3D and HDBox3D, we select the checkpoint that achieves the highest IoU@1 on the ScanNet validation set.
\section{Additional ablation studies}
\label{sec:suppl_ablation}

This section provides additional ablation studies on the expansion ratio across datasets (\cref{sec:suppl_expansion_ratio}), weight sharing across datasets (\cref{sec:suppl_weight_sharing}), and uniform high-resolution processing (\cref{sec:suppl_uniform_high_resolution}).

\subsection{Expansion ratio across datasets}
\label{sec:suppl_expansion_ratio}

The main paper reports the effect of the expansion ratio on ScanNet.
We further evaluate the effect of the expansion ratio on S3DIS, KITTI-360, and ScanNet++, with VSA applied to KITTI-360.

\begin{table}[t]
    \caption{Effect of the expansion ratio on S3DIS, KITTI-360, and ScanNet++. Decoder GFLOPs are averaged over clicks 2--10.}
    \label{tab:suppl_expansion_ratio}
    \centering
    \resizebox{\columnwidth}{!}{%
    \begin{tabular}{l|c|ccc|c}
        \toprule
        \multirow{2}{*}{\textbf{Test Dataset}} &
        \multirow{2}{*}{\textbf{\begin{tabular}[c]{@{}c@{}}Expansion \\ Ratio\end{tabular}}} &
        \multicolumn{3}{c|}{\textbf{IoU}} &
        \multirow{2}{*}{\textbf{GFLOPs $\downarrow$}} \\
        & & \textbf{@3} & \textbf{@5} & \textbf{@10} & \\
        \midrule

        \multirow{5}{*}{S3DIS}
              & 1.1 & 82.2 & 86.4 & 88.5 & 1.33 \\
              & 1.3 & 83.1 & \textbf{86.6} & 88.5 & 1.49 \\
              & 1.5 & 83.2 & 86.4 & 88.4 & 1.70 \\
              & 1.7 & 83.4 & \textbf{86.6} & \textbf{88.6} & 1.93 \\
              & 1.9 & \textbf{83.5} & 86.4 & 88.5 & 2.18 \\
        \midrule

        \multirow{5}{*}{KITTI-360}
              & 1.1 & 71.5 & \textbf{77.9} & 82.6 & 0.46 \\
              & 1.3 & 68.8 & 76.5 & 80.8 & 0.57 \\
              & 1.5 & 71.4 & 77.7 & 81.9 & 0.68 \\
              & 1.7 & \textbf{71.7} & 77.6 & 82.3 & 0.84 \\
              & 1.9 & 70.9 & 77.7 & \textbf{82.9} & 1.02 \\
        \midrule

        \multirow{5}{*}{ScanNet++}
              & 1.1 & 67.2 & 70.6 & 73.7 & 0.90 \\
              & 1.3 & 67.4 & 71.0 & 74.4 & 0.99 \\
              & 1.5 & 67.7 & \textbf{71.1} & 74.4 & 1.23 \\
              & 1.7 & \textbf{67.8} & \textbf{71.1} & \textbf{74.7} & 1.35 \\
              & 1.9 & 67.0 & 70.2 & 73.5 & 1.84 \\

        \bottomrule
    \end{tabular}}
\end{table}

As shown in \cref{tab:suppl_expansion_ratio}, the segmentation accuracy is relatively stable across different expansion ratios, while the computational cost increases as the ratio becomes larger.
Although the optimal ratio varies slightly across datasets and click settings, $r=1.7$ consistently achieves competitive performance, yielding the best or near-best results on all three datasets.

\subsection{Weight sharing across datasets}
\label{sec:suppl_weight_sharing}

\begin{table*}[t]
    \caption{Effect of weight sharing between coarse and fine stages.}
    \label{tab:suppl_weight_sharing}
    \centering
    \resizebox{1.7\columnwidth}{!}{%
    \begin{tabular}{l|c|c|cccccc|c}
        \toprule
        \multirow{2}{*}{\textbf{Test Dataset}} &
        \multirow{2}{*}{\textbf{U-Net}} &
        \multirow{2}{*}{\textbf{CSA Module}} &
        \multicolumn{6}{c|}{\textbf{IoU}} &
        \multirow{2}{*}{\textbf{\#Params}} \\
        & & &
        \textbf{@1} &
        \textbf{@2} &
        \textbf{@3} &
        \textbf{@5} &
        \textbf{@7} &
        \textbf{@10} & \\
        \midrule
        \multirow{3}{*}{S3DIS}
              & Separate & Separate & \textbf{69.8} & \textbf{79.7} & 85.0 & \textbf{89.7} & \textbf{91.6} & \textbf{93.0} & 27.77M \\
              & Shared & Separate & 69.6 & 79.6 & \textbf{85.2} & \textbf{89.7} & \textbf{91.6} & \textbf{93.0} & 16.81M \\
              & Shared & Shared & 69.1 & 79.0 & 84.4 & 89.3 & \textbf{91.6} & \textbf{93.0} & 13.89M \\
        \hline
        \multirow{3}{*}{KITTI-360}
                  & Separate & Separate & 54.0 & \textbf{64.3} & \textbf{71.9} & \textbf{79.8} & \textbf{83.2} & 85.8 & 27.77M \\
                  & Shared & Separate & 53.7 & 63.5 & 70.4 & 78.4 & 82.2 & 85.0 & 16.81M \\
                  & Shared & Shared & \textbf{54.8} & 63.4 & 70.7 & 78.8 & 82.8 & \textbf{85.9} & 13.89M \\
        \hline
        \multirow{3}{*}{ScanNet++}
                  & Separate & Separate & \textbf{60.4} & \textbf{67.7} & 72.0 & 76.8 & 79.3 & 81.4 & 27.77M \\
                  & Shared & Separate & 59.3 & 67.3 & \textbf{72.3} & \textbf{77.2} & \textbf{79.6} & \textbf{81.5} & 16.81M \\
                  & Shared & Shared & 58.8 & 66.8 & 71.5 & 76.3 & 79.0 & 81.0 & 13.89M \\
        \bottomrule
    \end{tabular}}
\end{table*}

\begin{table*}[t]
    \caption{Comparison with uniform high-resolution processing.}
    \label{tab:suppl_uniform_high_resolution}
    \centering
    \resizebox{2.0\columnwidth}{!}{%
    \begin{tabular}{l|l|cccccc|ccc|cc}
        \toprule
        \multirow{2}{*}{\textbf{Test Dataset}} &
        \multirow{2}{*}{\textbf{Method}} &
        \multicolumn{6}{c|}{\textbf{IoU}} &
        \multicolumn{3}{c|}{\textbf{GFLOPs}} &
        \multicolumn{2}{c}{\textbf{Peak Mem (MiB)}} \\
        & &
        \textbf{@1} &
        \textbf{@2} &
        \textbf{@3} &
        \textbf{@5} &
        \textbf{@7} &
        \textbf{@10} &
        \textbf{U-Net $\downarrow$} &
        \textbf{@1 $\downarrow$} &
        \textbf{@2--10 $\downarrow$} &
        \textbf{Avg. $\downarrow$} &
        \textbf{Max $\downarrow$} \\
        \midrule

        \multirow{2}{*}{ScanNet}
        & FastBox3D
        & \textbf{72.2} & \textbf{79.1} & \textbf{82.3} & 84.8 & 86.0 & 86.9
        & 40.018 & 67.711 & 5.632
        & 861.751 & 2084.521 \\

        & HDBox3D
        & 71.0 & 78.2 & 81.8 & \textbf{84.9} & \textbf{86.3} & \textbf{87.4}
        & 11.830 & 33.527 & 5.899
        & 387.678 & 1710.711 \\
        \midrule

        \multirow{2}{*}{S3DIS}
        & FastBox3D
        & \textbf{69.8} & \textbf{80.3} & \textbf{85.2} & 89.4 & 90.9 & 92.0
        & 27.193 & 60.777 & 7.364
        & 788.538 & 2642.050 \\

        & HDBox3D
        & 69.6 & 79.6 & \textbf{85.2} & \textbf{89.7} & \textbf{91.6} & \textbf{93.0}
        & 7.562 & 31.381 & 7.631
        & 370.468 & 2014.355 \\
        \midrule

        \multirow{2}{*}{KITTI-360}
        & FastBox3D$^*$
        & 49.4 & 62.2 & 69.9 & 78.2 & 82.0 & 84.7
        & 10.195 & 14.569 & 3.366
        & 237.895 & 962.820 \\

        & HDBox3D$^*$
        & \textbf{53.7} & \textbf{63.5} & \textbf{70.4} & \textbf{78.4} & \textbf{82.2} & \textbf{85.0}
        & 3.033 & 8.413 & 3.325
        & 209.554 & 870.514 \\
        \midrule

        \multirow{2}{*}{ScanNet++}
        & FastBox3D
        & 58.2 & 66.8 & 71.0 & 75.2 & 77.3 & 79.1
        & 108.219 & 216.284 & 6.664
        & 2619.718 & 6424.244 \\

        & HDBox3D
        & \textbf{59.3} & \textbf{67.3} & \textbf{72.3} & \textbf{77.2} & \textbf{79.6} & \textbf{81.5}
        & 29.182 & 72.357 & 5.072
        & 825.973 & 3967.258 \\

        \bottomrule
    \end{tabular}}
\end{table*}

The main paper reports the weight-sharing ablation on ScanNet. We additionally evaluate the three configurations on S3DIS, KITTI-360, and ScanNet++, using VSA for KITTI-360.

As shown in \cref{tab:suppl_weight_sharing}, sharing the sparse U-Net reduces the number of parameters while maintaining comparable segmentation accuracy across all three datasets.
Using a shared U-Net with separate CSA Modules provides a favorable balance between model size and accuracy, while additionally sharing the CSA Module further reduces the parameter count with only a small change in performance.

\subsection{Uniform high-resolution processing}
\label{sec:suppl_uniform_high_resolution}
To examine whether high-resolution training alone is sufficient without HDBox3D, we attempt to fine-tune Easy3D and FastBox3D at a voxel size of $0.025 \,\mathrm{m}$ using the same fine-tuning settings as HDBox3D.
Easy3D runs out of memory during training, even on four A100 GPUs with 80 GB of memory each, while FastBox3D can be successfully fine-tuned.

\cref{tab:suppl_uniform_high_resolution} compares HDBox3D with FastBox3D fine-tuned at a voxel size of $0.025 \,\mathrm{m}$. For KITTI-360, both methods are evaluated with VSA. In the GFLOPs columns, U-Net denotes the computation of the initial full-scene sparse U-Net encoding. FastBox3D encodes the entire scene at a voxel size of $0.025 \,\mathrm{m}$, whereas HDBox3D encodes the full scene at $0.05 \,\mathrm{m}$ and applies $0.025 \,\mathrm{m}$ encoding only to the localized fine region.
As shown in \cref{tab:suppl_uniform_high_resolution}, HDBox3D achieves segmentation accuracy comparable to or higher than that of high-resolution FastBox3D while requiring substantially less computation and memory.
\section{Additional qualitative results}
\label{sec:suppl_qualitative}

\begin{figure*}[t]
    \centering
    \includegraphics[width=\textwidth]{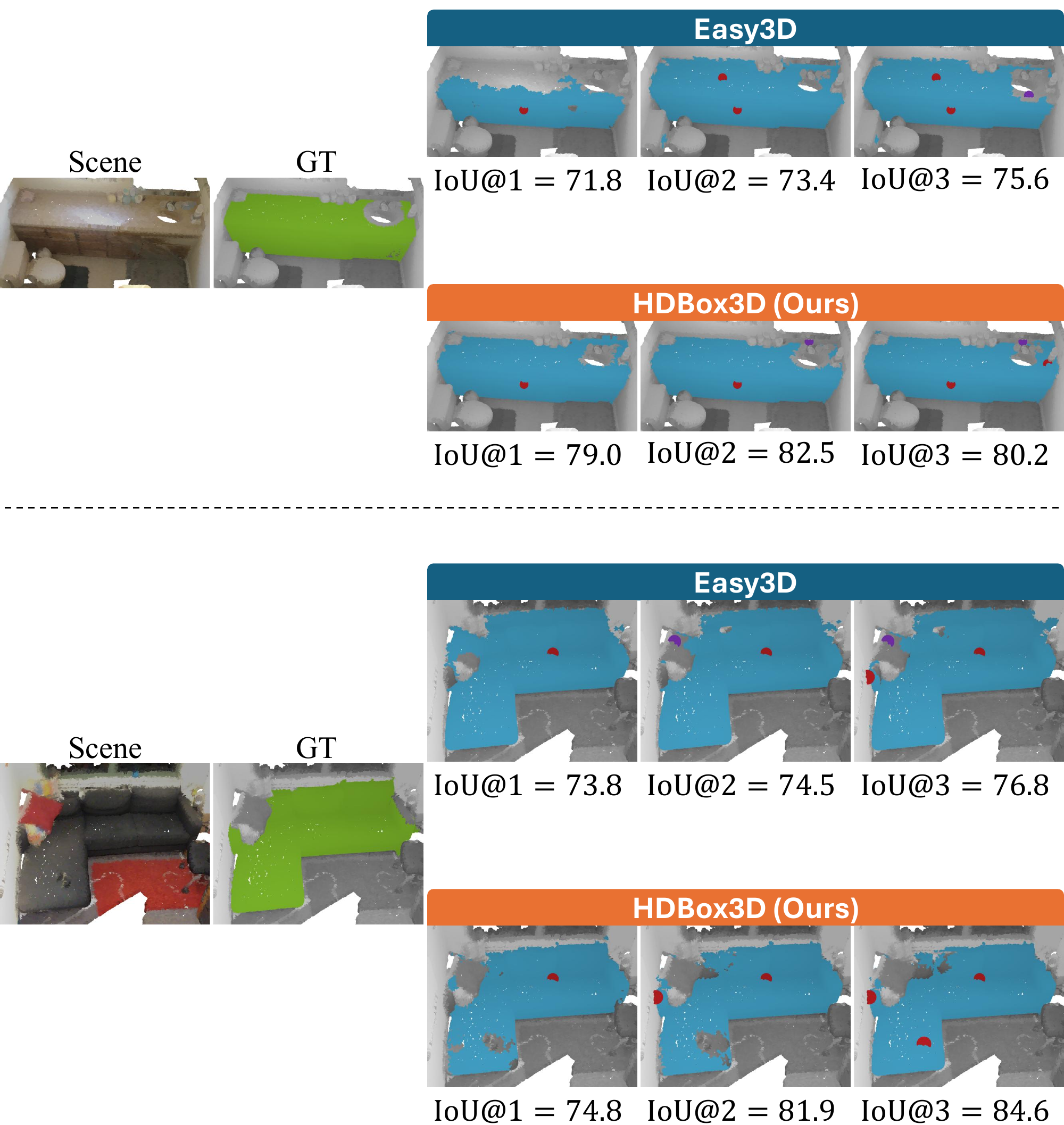}
    \caption{Qualitative comparison of Easy3D and HDBox3D on ScanNet.
    }
    \label{fig:sup_scannet}
\end{figure*}

\begin{figure*}[t]
    \centering
    \includegraphics[width=\textwidth]{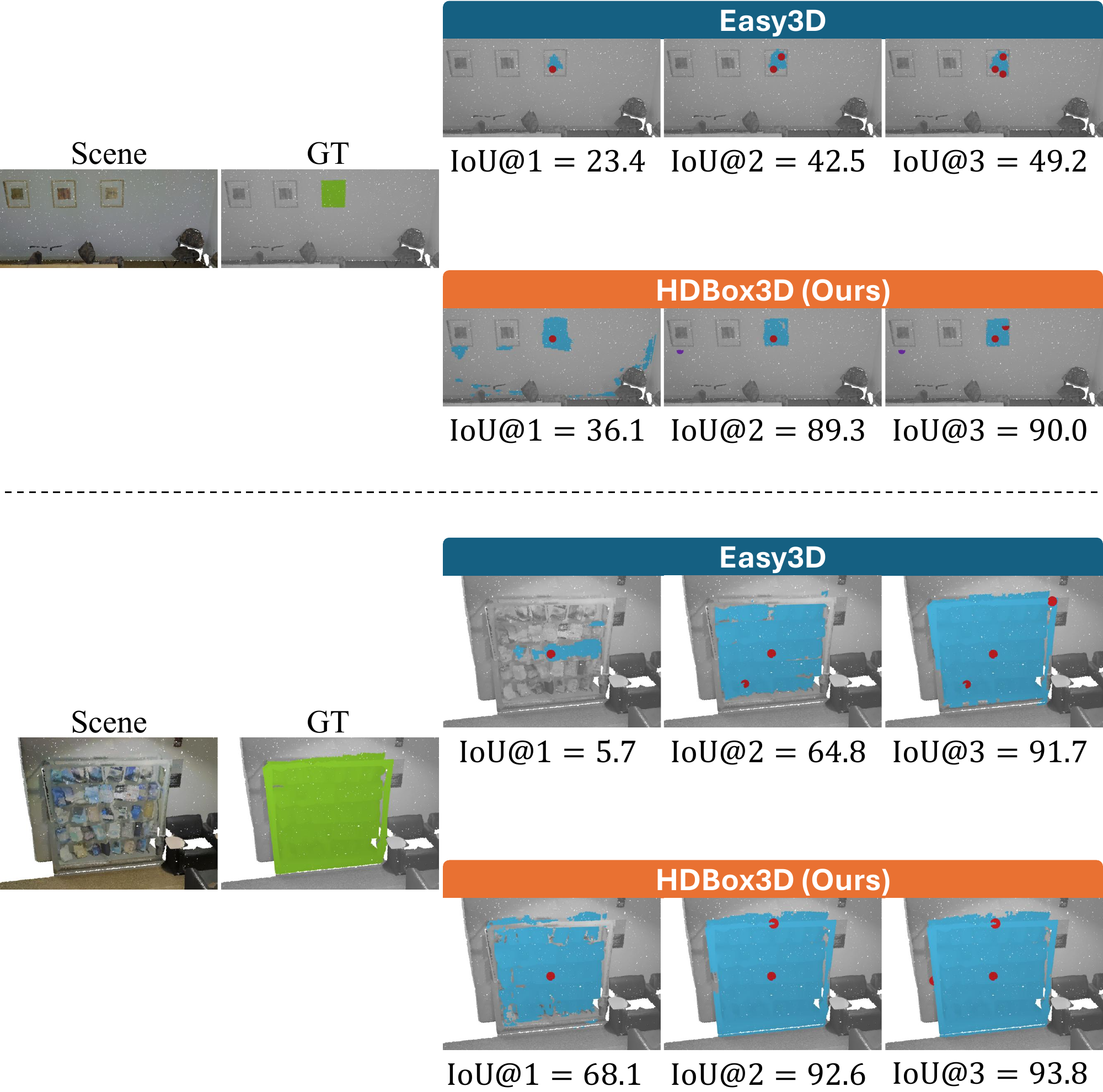}
    \caption{Qualitative comparison of Easy3D and HDBox3D on S3DIS.
    }
    \label{fig:sup_s3dis}
\end{figure*}

\begin{figure*}[t]
    \centering
    \includegraphics[width=\textwidth]{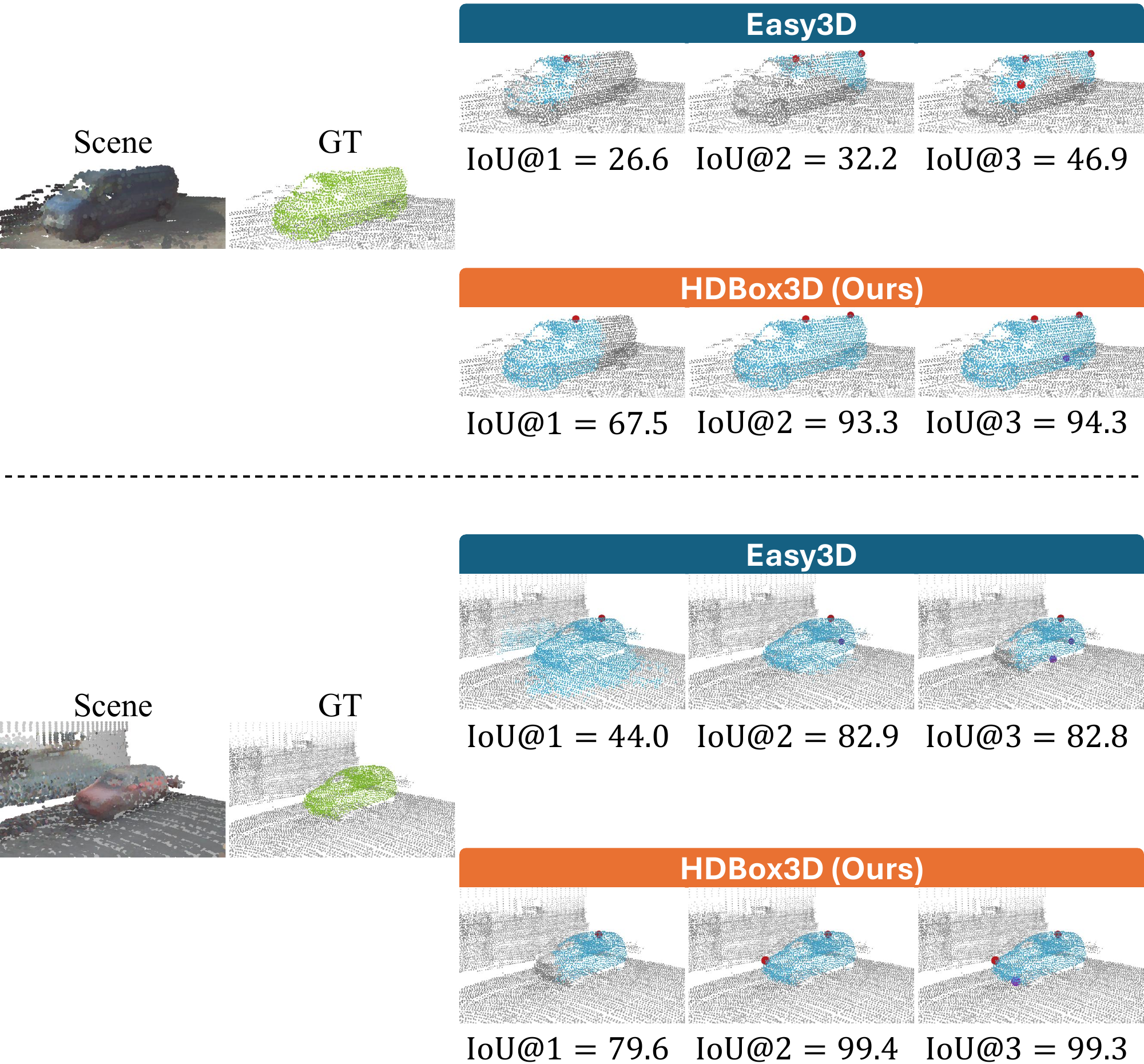}
    \caption{Qualitative comparison of Easy3D and HDBox3D on KITTI-360.
    }
    \label{fig:sup_kitti}
\end{figure*}

\begin{figure*}[t]
    \centering
    \includegraphics[width=\textwidth]{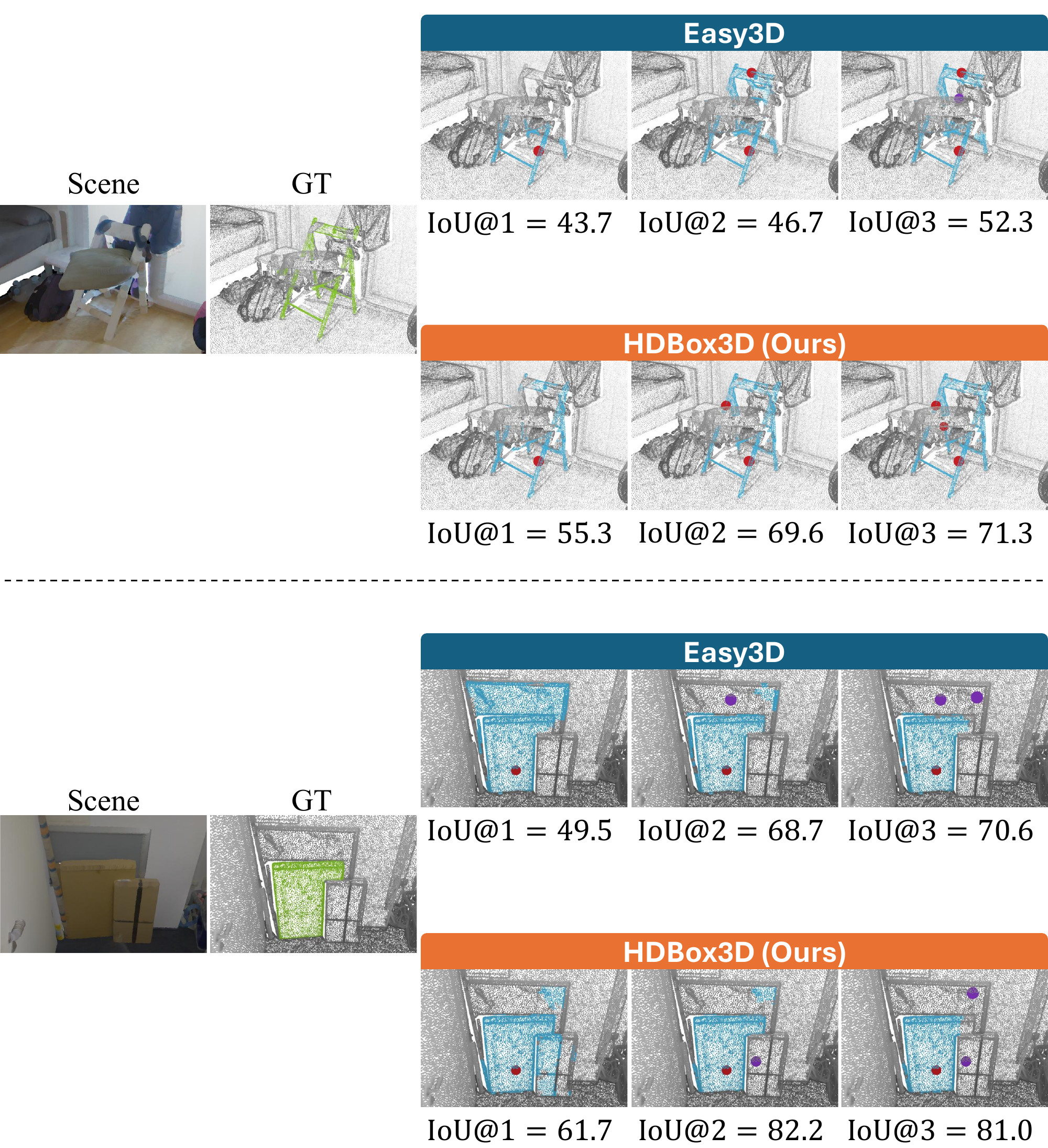}
    \caption{Qualitative comparison of Easy3D and HDBox3D on ScanNet++.
    }
    \label{fig:sup_scannetpp}
\end{figure*}

We provide additional qualitative comparisons between Easy3D and HDBox3D on ScanNet (\cref{fig:sup_scannet}), S3DIS (\cref{fig:sup_s3dis}), KITTI-360 (\cref{fig:sup_kitti}), and ScanNet++ (\cref{fig:sup_scannetpp}).
For KITTI-360, HDBox3D is evaluated with VSA, consistent with the qualitative evaluation in the main paper.

\section{Limitations}
\label{sec:suppl_limitations}
First, our methods still require full-scene processing for the first interaction because no prior prediction is available to localize the target object. Although this computation may be necessary to identify the target region from the entire scene, reducing the cost of the initial interaction remains an important direction for future work.

Second, our current framework performs segmentation on a voxelized representation, where all points assigned to the same voxel share the same prediction. This imposes a limit on the spatial resolution of the output and may be particularly restrictive for densely sampled point clouds such as ScanNet++, where fine geometric details can exist within a single voxel. While HDBox3D mitigates this issue through finer voxelization, an alternative direction would be to avoid voxelization and perform segmentation directly on points.

\end{document}